\documentclass[pmlr]{jmlr} % This is the PMLR single-column format

\usepackage{booktabs} % For better tables
\usepackage{graphicx} % For including figures
\usepackage{url}      % For URLs
\usepackage{amsmath}  % For mathematical notation
\usepackage{algorithm2e} % For algorithms (optional)

\jmlrvolume{TBD}
\jmlryear{2026}
\jmlrworkshop{Probabilistic Graphical Models (PGM)}

\title{Neuro-Causal Factor Analysis}

\author{\Name{Alex Markham} \Email{alex.markham@causal.dev}\\
  \addr Department of Mathematical Sciences and Pioneer Centre for SMARTbiomed\\
  University of Copenhagen, Denmark
  \AND
  \Name{Mingyu Liu}\\
  \addr Stern School of Business, New York University, USA
  \AND
  \Name{Bryon Aragam}\\
  \addr Booth School of Business, University of Chicago, USA
  \AND
  \Name{Liam Solus}\\
  \addr Department of Mathematics, KTH Royal Institute of Technology, Sweden
}
\editor{Gustau Camps-Valls, Manuele Leonelli and Gherardo Varando}

\setcitestyle{numbers}
\setcitestyle{square}

\usepackage[nameinlink]{cleveref}
\usepackage{tkz-graph}
\usepackage{mathtools}

\usepackage{enumitem}

\usepackage{tikz}
\usetikzlibrary{arrows.meta, positioning}

\newcommand{\edge}{\mathrel{\tikz[baseline=-0.5ex]{\draw[-] (0,0) -- (2.5ex,0);}}}

\tikzset{
  nodecirc/.style={
    circle,
    inner sep=0.06em,
    >=Stealth
  }
}

\newcommand{\insertsimulated}{%
	\begin{tabular}{rcccccc}
		\toprule
		                   & \textbf{MSE $\downarrow$} & \textbf{MCC $\uparrow$} & \textbf{dCor $\uparrow$} & \textbf{SFD $\downarrow$} & \textbf{Time (s)}  & \textbf{\# Params} \\
		\midrule
		\textbf{FA}        & 13.673 $\pm$ 7.358        & 0.459 $\pm$ 0.051       & 0.445 $\pm$ 0.036        & 508 $\pm$ 2               & 0.655 $\pm$ 0.132  & 120 $\pm$ 0        \\
		\textbf{VAE}       & 0.251 $\pm$ 0.027         & 0.756 $\pm$ 0.055       & 0.768 $\pm$ 0.017        & 508 $\pm$ 2               & 52.644 $\pm$ 7.254 & 7350 $\pm$ 0       \\
		\textbf{LG-minMCM} & 0.965 $\pm$ 0.076         & 0.916 $\pm$ 0.016       & 0.908 $\pm$ 0.016        & 7 $\pm$ 6                 & 0.061 $\pm$ 0.019  & 35 $\pm$ 1         \\
		\textbf{NCFA}      & 0.928 $\pm$ 0.078         & 0.898 $\pm$ 0.023       & 0.897 $\pm$ 0.018        & 5 $\pm$ 5                 & 44.357 $\pm$ 9.329 & 6300 $\pm$ 0       \\
		\bottomrule
	\end{tabular}%
}

\newcommand{\insertcausalchamber}{%
	\begin{tabular}{rcccccc}
		\toprule
		                   & \textbf{MSE $\downarrow$} & \textbf{MCC $\uparrow$} & \textbf{dCor $\uparrow$} & \textbf{SFD $\downarrow$} & \textbf{Time (s)}  & \textbf{\# Params} \\
		\midrule
		\textbf{FA}        & 0.334 $\pm$ 0.048         & 0.346 $\pm$ 0.045       & 0.565 $\pm$ 0.065        & 169 $\pm$ 0               & 0.202 $\pm$ 0.048  & 63 $\pm$ 0         \\
		\textbf{VAE}       & 0.201 $\pm$ 0.004         & 0.503 $\pm$ 0.019       & 0.697 $\pm$ 0.002        & 169 $\pm$ 0               & 27.186 $\pm$ 6.171 & 6321 $\pm$ 0       \\
		\textbf{LG-minMCM} & 0.566 $\pm$ 0.138         & 0.679 $\pm$ 0.014       & 0.570 $\pm$ 0.046        & 3 $\pm$ 2                 & 0.051 $\pm$ 0.023  & 24 $\pm$ 1         \\
		\textbf{NCFA}      & 0.209 $\pm$ 0.008         & 0.885 $\pm$ 0.003       & 0.694 $\pm$ 0.003        & 10 $\pm$ 0                & 20.725 $\pm$ 4.986 & 5745 $\pm$ 0       \\
		\bottomrule
	\end{tabular}%
}

\SetKwInOut{Parameter}{Parameter}

\newcommand\independent{\protect\mathpalette{\protect\independenT}{\perp}}
\def\independenT#1#2{\mathrel{\rlap{$#1#2$}\mkern2mu{#1#2}}}

\begin{document}
\maketitle

\begin{abstract}
	Factor analysis (FA) is a statistical method for explaining how mutually dependent observed variables can be represented in terms of mutually independent latent factors, and it is widely used in the psychological, biological, and physical sciences.
	We revisit this classic method from the perspective of recent advances in causal structure learning and deep generative models, introducing a framework for \emph{Neuro-Causal Factor Analysis (NCFA)}.
	Our approach is fully nonparametric: it learns a directed graph between latent and observed variables, and then fits a deep generative model constrained to respect the Markov factorization of the graph.
	Empirically, on synthetic and real data, NCFA attains better reconstruction error compared to standard FA and better latent distribution recovery compared to a standard variational autoencoder, all with the advantages of sparser architecture, lower model complexity, and causal interpretability.
\end{abstract}


\section{Introduction}
\label{sec:introduction}

Since its development over a century ago, factor analysis (FA) \citep{spearman1904general} has been applied across scientific fields, including genomics, computational biology \citep{pournara2007factor, velten2022identifying}, economics \citep{forni1998let,ludvigson2007empirical}, sociology \citep{bollen2012instrumental} and many others.
The goal of FA is to offer explanations of variability among dependent observables via (potentially) fewer latent variables that capture the joint dependence between the observables.
For the sake of identifiability, it is common to assume linearity, although in practice it is well-known that many problems exhibit complex nonlinear latent structures.
With the rise of nonparametric deep generative models that allow representing highly nonlinear relationships between dependent observables, one might hope to combine the best of both worlds.

Moreover, within applications such as those listed above, FA is considered useful because the learned factors (latents) may offer an interpretation of relevant observed correlations.
Many applied FA studies provide an interpretation of the learned factors based on the observed variables whose joint correlation they encode.
A natural tendency when trying to interpret these factors is to assume they reflect possible common causes linking observed variables.
However, the models used in such studies are not necessarily built with causality in mind.
Collectively, these considerations purport a need for a framework for nonlinear \emph{causal} factor analysis that combines identifiability with flexibility through the use of modern advances in deep generative models and causality.

To this end, we propose \emph{Neuro-Causal Factor Analysis (NCFA)}, augmenting classic FA on both fronts by leveraging advancements of the last few decades, including (i) causal discovery \citep{spirtes2000causation,pearl2009causality} and (ii) deep generative models, such as variational autoencoders (VAEs) \citep{kingma2013auto,rezende2014stochastic}.
To formalize this combination of ideas and apply it to settings where FA is typically invoked, we consider causal models that directly abide by \emph{Reichenbach's common cause principle} \citep[p.~157]{reichenbach1956direction}: dependent variables in a system that do not share a direct causal relation should be explained by the existence of unobserved common causes which when conditioned upon render them independent.
In particular, NCFA is applicable to problems where one can assume that the observed (or \emph{measurement}) variables are rendered mutually independent when conditioning on a set of unobserved latent variables, which may be interpreted as \emph{causally justifiable} factors from the FA perspective.

\paragraph{Contributions}

\begin{figure}[t]
	\centering
	\includegraphics[width=\linewidth]{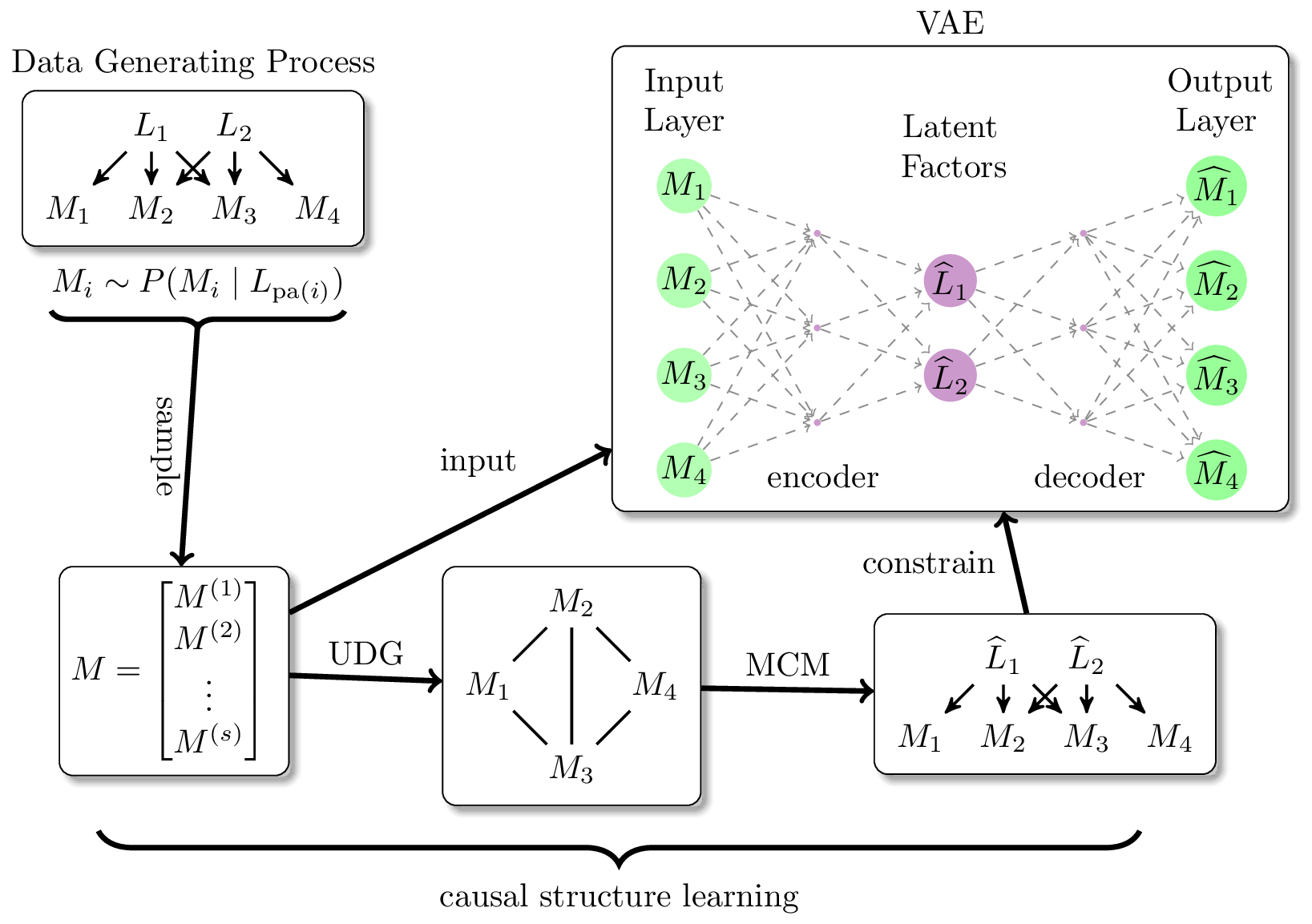}
	\caption{Learning a Neuro-Causal Factor Model.
	}
	\label{fig:pipeline}
\end{figure}

Our main contribution is the NCFA framework (\Cref{fig:pipeline}) for causally interpretable, identifiable FA models with the flexibility afforded by deep generative models.
Unlike recent work (\Cref{sec:related-work}) that seeks to combine deep learning with causal models, NCFA is model-free and fully nonparametric, relying only on simple graphical assumptions.
This is in contrast to approaches that rely on functional restrictions such as additive noise, post-nonlinearity, or auxiliary information.
Instead, we ensure identifiability through the graphical properties of the marginal independence structure induced by a latent factor model, which has not been previously leveraged for deep causal modeling.
These assumptions arise in practical applications \citep[e.g.,][]{bing2019latent,bing2022essential,rahimikollu2024slide} and are weaker than faithfulness.
As a result, we obtain identifiability without imposing parametrization, functional relations, or conditions on data type (e.g., continuous, discrete, or mixed).
Overall, NCFA provides a model-free perspective on latent causal models.
Specifically:

\begin{enumerate}[itemsep=4pt, topsep=4pt]
	\item NCFA (\Cref{alg:ncfa}) does not assume the underlying structure is known (i.e., it is learned entirely from data), allows flexible estimation of the latent space with deep generative models, and comes with fully nonparametric graph identifiability guarantees (\Cref{thm: ident}; see \Cref{rem:latent-ident} for the distinction from full identifiability of the latent distribution).
	      Unlike traditional causal discovery methods, NCFA relies only on \emph{marginal} independence testing rather than general conditional independence testing, which is notoriously difficult \citep{neykov2021minimax,canonne2018testing,shah2020hardness}.
	      As a result, the framework applies to any combination of discrete, continuous, or mixed variables as long as marginal dependence can be consistently tested.
	      The advantages of this approach are illustrated with concrete examples (\Cref{appsec: identifiability}) that cannot be captured by existing approaches.

	\item We introduce the notion of \emph{latent degrees of freedom} (\Cref{sec:ncfa-models}), where additional representational capacity is achieved by giving each causal variable its own factorial prior.
	      This separates neural network architectural choices from defining properties of the underlying causal graph.

	\item We demonstrate (\Cref{sec:appl}) on both synthetic and real data that NCFA injects generative models with interpretable structure and latent representations without significantly reducing reconstruction quality compared to unstructured generative models.

	\item We provide an open source Python implementation as part of the \texttt{medil} package\footnote{\url{https://medil.causal.dev}}, along with a Snakemake workflow for reproducing all results\footnote{\url{https://gitlab.com/alex-markham/medil/-/tree/v2.0.0/src/expt}}.
\end{enumerate}

\section{Comparison to related work}
\label{sec:related-work}

We divide the vast amount of related work into three areas: factor analysis (\Cref{sec:factor-analysis}), latent causal discovery (\Cref{sec:latent-causal-models}), and deep generative models (\Cref{sec:deep-gener-models}).
Before describing each in more detail in its respective subsection, we first summarize their differing motivations and methods and provide a comparison to our proposed NCFA.
A comprehensive comparison with recent related work on deep causal models can be found in \Cref{app:detailedcomp}.

Factor analysis focuses on modeling measurement variables in terms of underlying factors (which can be interpreted as sources), focusing on model simplicity and interpretability, generally by assuming linear relations and jointly Gaussian random variables.
Latent causal models focus on more detailed causal structure, not being limited to measurement variables and their latent sources, resulting in extremely interpretable models, but often at the expense of (arguably) strong, untestable assumptions like faithfulness.
Deep generative models focus on learning as accurate a black box model as possible, optimizing a highly overparameterized and nonlinear model to still achieve generalizability.
Although the interpretation of deep generative models as nonlinear factor analysis is standard in the literature \citep[e.g.,][]{roweis1999unifying,murphy2022probabilistic,goodfellow2016deep}, the additional dimensions of causality and identifiability are new to our approach.
NCFA offers a unifying perspective on structured representation learning, incorporating the strengths of each of these approaches.

Like FA, we focus on modeling measurement variables in terms of their underlying sources; however, NCFA identifies these sources and their structural connections to the measurements through explicit latent causal structure learning, which is easier and requires weaker assumptions by focusing on sources cause measurements instead of more detailed intermediate causal structure.
Furthermore, the source distributions and their corresponding functional relations to the measurements are estimated using a generative model whose architecture is constrained to respect the learned causal structure, gaining some of the expressiveness of deep generative models but regularized to maintain causal interpretability and generalizability.
Hence, NCFA is motivated by the simplicity of FA, the causal interpretability of latent causal models, and the expressive power of deep generative models.

\subsection{Factor analysis}
\label{sec:factor-analysis}

We now give a brief introduction to FA, focusing on the key terms and mathematical ideas that we connect to latent causal discovery and deep generative models.
For a more in-depth introduction, see \citet{mulaik2009foundations}.

\begin{definition}
	\label{def:linear-factor-model}
	A \emph{factor model} is defined by \(\mathbf{M} = \mathbf{L}W + \boldsymbol{\epsilon}\).
	A random (row) vector \(\mathbf{M} \sim \mathcal{N}(\mathbf{0}, \mathbf{\Sigma})\) consisting of \(n\) \emph{measurement variables} is represented as a linear transformation of a standard jointly normal random vector \(\mathbf{L} \sim \mathcal{N}(\mathbf{0}, I_K)\) of \(K < n\) \emph{latent factors} via \emph{factor loading weights} \(W\in\mathbb{R}^{K\times n}\) plus a jointly normal random vector of $n$ \emph{error terms} \(\boldsymbol{\epsilon} \sim \mathcal{N}(\mathbf{0}, D)\), where \(D\in\mathbb{R}_+^{n\times n}\) is a diagonal matrix.
\end{definition}

Given a sample \(M\in \mathbb{R}^{s, n} \sim \mathbf{M}\) and assuming that \(\mathbf{L}\) and \(\boldsymbol{\epsilon}\) are probabilistically independent, the factor model can be estimated \citep{adachi2019factor} from the empirical covariance matrix \(\widehat{\Sigma} = \frac{1}{s}M^\top M\) by finding \(\widehat{W}\) and \(\widehat{D}\) that minimize the squared Frobenius norm \(\lVert \widehat{\Sigma} - \widehat{W}^\top \widehat{W} - \widehat{D}\rVert^2_F\).
Such a solution is unique only up to orthogonal transformations of \(\widehat{W}\), and so without further (e.g., in our case, causal) assumptions, finding a solution does not always warrant a meaningful interpretation of the resulting factor model.
This unidentifiability poses a problem in \emph{exploratory} FA, where there is no prior knowledge about \(\widehat{\Sigma}, \widehat{W}\) or \(\widehat{D}\), but less so in \emph{confirmatory} FA, where experts incorporate domain knowledge to constrain and interpret solutions as well as test specific hypotheses.

Additionally, there are possibilities for either restricting or relaxing the FA model, including closely related methods like PCA \citep{pearson1901liii,hotelling1933analysis,jolliffe2002principal}, ICA \citep{comon1994independent,hyvarinen2000independent}, and many others beyond our scope.
Notably, compared to other related work, sparse FA \citep{ning2011sparse,trendafilov2017sparse,yamamoto2017graphical}, which penalizes \(\widehat{W}\) according to the number of nonzero entries, produces solutions more closely related to those we find with NCFA.
The two main differences between sparse FA and NCFA are that (i) rather than explicitly penalizing the solution to encourage sparsity, NCFA simply learns a causal structure that exhibits a structure typically sought in sparse FA, and (ii) like most FA methods, sparse FA still assumes linearity and Gaussianity, whereas NCFA can be highly nonlinear and nonparametric.
More recently, \citet{kim23structure} proposed a method similar to the one proposed here, restricted to linear models.
Their identifiability results rely on their Condition 4.2, and are a linear special case of our \Cref{cor:ident}.

\subsection{Latent causal models}
\label{sec:latent-causal-models}

Graphical causal modeling \citep{spirtes2000causation,pearl2009causality} focuses on learning a directed acyclic graph (DAG) representation of the causal relations among variables.
This typically requires a strengthening of the common cause principle (into what is sometimes called the causal Markov assumption), which additionally assumes causal sufficiency, i.e., that there are no latent variables, and hence that all probabilistic dependences among the observed variables are due to causal relations among them.
Methods for learning latent causal models have classically focused on learning DAG-like structure (using mixed instead of only directed graphs) among the observed variables to the extent allowed by confounding latent variables, exemplified by algorithms such as FCI \citep{spirtes2000causation,colombo2012learning} and IC \citep{pearl1995theory}, which relax the Causal Markov Assumption.
We also mention early work on this problem by \citep{martin1995discrete,friedman1997learning,elidan2000discovering}.
In contrast, research on causal measurement models \citep{silva2003} is more closely related to the goal of FA, in that it too focuses on factor-measurement relations.
Recently, there has been a surge of interest in these models, with advances leveraging additive noise models \citep{maeda2021causal,NEURIPS2022_05b63fa0,huang2022latent,xie2022identification,ashman2022causal}, independent mechanisms \citep{gresele2021independent}, weak supervision \citep{liu2022identifying,brehmer2022weakly}, and interventions \citep{chalupka2014visual,chalupka2017causal,ahuja2022interventional,seigal2022linear,varici2023score}.

\subsection{Structured deep generative models}
\label{sec:deep-gener-models}

The past decade has seen a flurry of work on training large-scale deep latent variable models, fueled by advances in variational inference and deep learning \citep[e.g.][]{larochelle2011neural,kingma2013auto,rezende2014stochastic,dinh2014nice,goodfellow2014generative,rezende2015variational,sohl2015deep}.
More recently, there has been a trend towards \emph{structured} latent spaces, such as hierarchical, graphical, causal, and disentangled structures.
Conceptually, NCFA provides a theoretically principled approach to automatically \emph{learning} latent structure from data in a way that is causally meaningful.
The related work here needs to be divided into two categories: \emph{known} (e.g., from prior knowledge) vs.
\emph{learned} latent structure.
These can be further divided into \emph{non-causal} vs.
\emph{causal} approaches.
Given that our main contribution is \emph{learned causal structure}, we will focus the discussion on the latter:
For causal structure, \emph{identifiability} becomes crucial, as it is well-known that nonparametric latent variable models are unidentifiable in general \citep{hyvarinen1999nonlinear,locatello2019challenging}.

\paragraph{Known structure}
Early work looked at incorporating known structure into generative models, such as autoregressive, graphical, and hierarchical structure \citep{germain2015made,johnson2016composing,sonderby2016ladder,webb2018faithful,weilbach2020structured,ding2021gans,moutonintegrating}.
This was later translated into known \emph{causal} structure \citep{kocaoglu2018causalgan}.

\paragraph{Learned structure}
When the latent structure is unknown, several techniques have been developed to learn useful (not necessarily causal) structure from data \citep{li2018learning,he2019variational,wehenkel2021graphical,kivva2022identifiability,moran2023identifiable}.
More recently, based on growing interest in disentangled \citep{bengio2013deep} and/or causal \citep{scholkopf2021toward} representation learning, methods that learn causal structure have been developed \citep{moraffah2020causal,yang2021causalvae,ashman2022causal,shen2022weakly,pmlr-v202-kaltenpoth23a}.
\citet{subramanian2022learning} assumes a linear Gaussian additive noise model, whereas \citet{moraffah2020causal} uses GANs.
Unlike NCFA, neither \citet{moraffah2020causal} nor \citet{subramanian2022learning} come with identifiability guarantees.
The remaining papers (detailed in \Cref{app:detailedcomp}) all rely on functional restrictions (e.g., additive noise models) or auxiliary information (e.g., as in iVAE) to prove identifiability.
In contrast to this existing work, NCFA admits nonparametric identifiability guarantees without additional labels or specifying a particular parametric or functional form (see \Cref{sec:identifiability}).

\section{NCFA models}
\label{sec:ncfa-models}

Consider a collection of jointly distributed measurement variables $(M_1,\ldots, M_n)$ for which we assume that all dependences are explained by the existence of latent common causes of the measured variables, i.e., that no $M_i$ and $M_j$ share a direct causal relation.
If we were able to observe these latent confounders and condition upon them, $M_1,\ldots, M_n$ would become mutually independent.
Since every $M_i$ has out-degree $0$, any $M_k$ lying on a path between two other measurements must itself be a collider, and any two dependent measurement variables must share a latent parent;
consequently the unconditional (in)dependencies among $M_1,\ldots, M_n$ already fully determine their conditional (in)dependencies \citep[Proposition~6]{markham20mcm}.
Hence, the only causal structure encoded via conditional independence in the observed distribution is contained in their marginal independence structure, which can be represented as an undirected graph:
\begin{definition}
	\label[definition]{def: UDG}
	The \emph{unconditional dependence graph} (UDG) for the jointly distributed random variables $(M_1,\ldots, M_n)$ is the undirected graph $\mathcal{U}$ with node set $[n] = \{1,\ldots, n\}$ and edge set \({E = \{i \edge j : M_i\not\independent M_j\}}\).
\end{definition}

To recover a causal interpretation of the relations that hold among the measurement variables, we extend a UDG graph to a \emph{(minimum) MCM\footnote{These were originally defined in the context of \emph{measurement dependence inducing latent (MeDIL) causal models} \citep{markham20mcm}.
		A summary of this theory is given in \Cref{appsec: medil models}.
	} graph}.
Following the principle of Occam's Razor, we would like to explain the observed dependences in $(M_1,\ldots, M_n)$ in the simplest possible way, i.e., using the fewest possible latents to serve as the common causes of the measurement variables that exhibit dependence.
To do so, we identify a \emph{minimum edge clique cover} (ECC) of the UDG $\mathcal{U}$, which is a collection $\mathcal{C} = \{C_1,\ldots, C_K\}$ of \emph{cliques} (i.e., complete subgraphs of $\mathcal{U}$) such that for every $i \edge j \in E$ the pair $i,j$ is contained in at least one clique in $\mathcal{C}$ and there exists no set of cliques with this property that has cardinality smaller than $\lvert\mathcal{C} \rvert$:
\begin{definition}
	\sloppy
	\label[definition]{def: mcm graph}
	Let $\mathcal{U}$ be an undirected graph with minimum edge clique cover $\mathcal{C} = \{C_1,\ldots, C_K\}$.
	The \emph{(minimum) MCM graph} $\mathcal{G}$ for $\mathcal{U}$ and $\mathcal{C}$ is the DAG with vertices $[n]\cup L$ where $L = \{l_1,\ldots, l_K\}$ and edge set \({E = \{l_i \rightarrow j : j\in C_i,\ i \in [K]\}}\).
	We call $\lvert L \rvert$ the number of \emph{causal degrees of freedom} of the model.
\end{definition}

Since we assumed all marginal dependencies in $(M_1,\ldots, M_n)$ are explainable by latent common causes, then the observed distribution $(M_1,\ldots, M_n)$ is realizable as the marginal distribution of $(M_1,\ldots, M_n)$ in the joint distribution $(M_1,\ldots, M_n, L_1, \ldots, L_K)$ that is Markov to the DAG $\mathcal{G}$, where $L_i$ is the random variable represented by the node $l_i$ in $\mathcal{G}$.
From a factor analysis perspective, the latents $L_1,\ldots, L_K$ are the factors to be inferred.

\tikzstyle{VertexStyle} = []
\SetGraphUnit{2.2}
\begin{figure}[t]
	\centering

	% ---------- Subfigure (1) ----------
	\subfigure[\(\mathcal{G}\)]{%
		\label{fig:medil graphs:G}%
		\begin{tikzpicture}[scale=0.6, baseline=(current bounding box.center)]
			\tikzstyle{EdgeStyle} = [->,>=stealth']
			\begin{scope}[execute at begin node=$, execute at end node=$]
				\SetGraphUnit{2.2}
				\Vertex{1} \EA(1){2} \EA(2){3} \EA(3){4}
				\NO(2){l_1} \EA(l_1){l_2}
				\Edges(l_1, 1) \Edges(l_1, 2) \Edges(l_1, 3)
				\Edges(l_2, 4) \Edges(l_2, 2) \Edges(l_2, 3)
			\end{scope}
		\end{tikzpicture}%
	}\qquad
	% ---------- Subfigure (2) ----------
	\subfigure[\(\mathcal{U}\)]{%
		\label{fig:medil graphs:U}%
		\begin{tikzpicture}[scale=0.4, baseline=(current bounding box.center)]
			\tikzstyle{EdgeStyle} = []
			\begin{scope}[execute at begin node=$, execute at end node=$]
				\SetGraphUnit{2.2}
				\Vertex[empty]{a} \EA(a){4} \NO(a){2} \SO(a){3} \WE(a){1}
				\Edges(1, 2, 4, 3, 1) \Edges(2, 3)
			\end{scope}
		\end{tikzpicture}%
	}\qquad
	% ---------- Subfigure (3) ----------
	\subfigure[\(\widetilde{\mathcal{G}}\)]{%
		\label{fig:medil graphs:ncfa}%
		\begin{tikzpicture}[scale=0.6, baseline=(current bounding box.center)]
			\tikzstyle{EdgeStyle} = [->,>=stealth']
			\begin{scope}[execute at begin node=$, execute at end node=$]
				\SetGraphUnit{2.2}
				\Vertex{1} \EA(1){2} \EA(2){3} \EA(3){4}
				\NO[L=\ell_{1,1}](1){L_1} \EA[L=\ell_{1,2}](L_1){L_2}
				\EA[L=\ell_{2,1}](L_2){L_3} \EA[L=\ell_{2,2}](L_3){L_4}
				\Edges(L_1, 1) \Edges(L_1, 2) \Edges(L_1, 3)
				\Edges(L_2, 1) \Edges(L_2, 2) \Edges(L_2, 3)
				\Edges(L_3, 4) \Edges(L_3, 2) \Edges(L_3, 3)
				\Edges(L_4, 4) \Edges(L_4, 2) \Edges(L_4, 3)
			\end{scope}
		\end{tikzpicture}%
	}
	\caption{\emph{(a)} shows an example minimum MCM graph with causal degrees of freedom \(|L|=2\), with corresponding UDG in \emph{(b)}, and example NCFA-graph with latent degrees of freedom \(\lambda=4\) shown in \emph{(c)}.
		Note that all three graphs encode the exact same set of (marginal) independencies among the measurement variables.
	}
	\label{fig:medil graphs}
\end{figure}

The minimum MCM graph defines a putative causal graph that respects the independence structure of $(M_1,\ldots, M_n)$, and our goal is to learn the associated latent representations from data using a deep generative model.
Consider the DAG $\mathcal{G}$ depicted in \Cref{fig:medil graphs:G} with two latents.
A na\"ive approach would be to design a standard generative model, such as a variational autoencoder (VAE), such that the decoder respects the Markov properties implied by $\mathcal{G}$, however, it is unlikely that any generative model trained with a two-dimensional latent space will be able to represent the measurement variables accurately.
The difficulty is that although the \emph{true} causal structure involves only two latent variables, exactly fitting such a model is very difficult in practice.
Thus, there is a tension between expressive capacity and respecting the causal structure.

We overcome this difficulty by replacing each causal latent with an overparameterized, factorial prior.
The virtue of overparameterization is well-documented in the literature \citep{radhakrishnan2020overparameterized,buhai2020empirical}; in our setting this has the effect of increasing representational capacity without breaking the Markov structure encoded in $\mathcal{G}$.
Formally, given a minimum MCM graph $\mathcal{G} = \langle [n] \cup L, E\rangle$, we replace each $l_i$ with a set of independent latent nodes $\mathcal{L}_i = \{\ell_{i,1},\ldots, \ell_{i,k_i}\}$, for some $k_i\geq 1$, each with the same connectivity (i.e., children) as $l_i$.
Thus, all told, we distribute $ \lambda = \sum_{i \in [K]}k_i$ latents across the cliques, a hyperparameter called the \emph{latent degrees of freedom}.
It is easy to check that no matter how the $\lambda$ latent degrees of freedom are distributed, the resulting DAG has the same independence structure over the measurement variables as $\mathcal{G}$.
This provides a rigorous device for increasing complexity without affecting the causal structure, and moreover, $\lambda$ is a flexible tuning hyperparameter that can be set arbitrarily large in practice, resulting in potentially overparameterized models.
We call the resulting graph a \emph{NCFA-graph} of $\mathcal{G}$ with $\lambda$ latent degrees of freedom:

\begin{definition}
	\label[definition]{def: NCFA graph}
	Let $\mathcal{G}$ be a minimum MCM graph for the UDG $\mathcal{U} = \langle [n], E\rangle$ and the minimum edge clique cover $\mathcal{C} = \{C_1,\ldots, C_k\}$ of $\mathcal{U}$.
	A \emph{NCFA graph} of $\mathcal{G}$ with $\lambda$ latent degrees of freedom is a graph $\widetilde{\mathcal{G}}$ with node set $[n]\cup \widetilde{\mathcal{L}}$ and edge set $\widetilde{E}$, where \(\widetilde{\mathcal{L}} = \mathcal{L}_1\cup \cdots \cup \mathcal{L}_k \) for \(\mathcal{L}_i = \{\ell_{i,1},\ldots,\ell_{i,k_i}\}\) with \(k_i\geq 1 \ \ \forall i\in[k],\) and \({\widetilde{E} = \{ \ell_{i,m}\rightarrow j : j\in C_i,\ m \in k_i,\ i\in[k]\}}\).
\end{definition}

\Cref{fig:medil graphs} shows an example of the three graph types considered so far in \Cref{def: UDG,def: mcm graph,def: NCFA graph}.
Now we shift focus from graphs to models.

Each node $\ell_{i,m}$ represents a latent variable $Z_{i,m}$.
Since the latent nodes in $\mathcal{L}_i$ all have the same connectivity as the single latent $l_i$, their joint distribution $f(L_i) = \prod_{m=1}^{k_i}f(Z_{i,m})$ represents the common cause of the measurement variables corresponding to the nodes in $C_i$, which was previously only represented by $l_i$ in $\mathcal{G}$.
The factors to be inferred from a factor analysis perspective are now the random vectors $L_1, \ldots, L_K$ with $L_i = (Z_{i,1},\ldots,Z_{i,k_i})$, which still have the causal interpretation afforded by the minimum MCM graph.
However, the multiple latents provide us flexibility to model the effects of the causal factors.

\begin{definition}
	\label{def: NCFA model}
	A \emph{NCFA model} is a joint distribution $(M_1,\ldots,M_n)$ for which there is an NCFA-graph $\widetilde{\mathcal{G}} = \langle [n]\cup \widetilde{\mathcal{L}}, \widetilde{E}\rangle$ and functions $f_1,\ldots, f_n$ for which $M_i \coloneqq f_i(\mathrm{pa}_Z(i), \epsilon_i)$ for all $i\in[n]$, where \(\mathrm{pa}_Z(i) \coloneqq \{Z_{j,m} \mathop{:} \ell_{j,m} \in \textrm{pa}_{\widetilde{G}}(i)\}\), and the measurement errors \(\epsilon_1,\ldots,\epsilon_n\) are jointly independent of each other and of \(Z\).
\end{definition}

When modeling a distribution via an NCFA model, the functions $f_i$ are treated as unknowns to be inferred via a deep generative model.
In our implementation, we used VAEs, however, other generative models can be used instead instead.
The encoder maps the observations into the latent space as the joint posterior distribution $f( Z \mid M_1,\ldots,M_n)$ where $Z$ is the random vector that collects the $Z_{j,m}$, and the decoder maps latents to observations according to \[ f(M_1,\ldots,M_n \mid Z) = \prod_{i=1}^{n}f(M_i \mid \mathrm{pa}_Z(i)).
\]
The joint distribution of the latent space is $f( Z ) = \prod_{i=1}^Kf(L_i)$; i.e., it is a product of the (joint) distributions we have specified to represent each of the latent common causes in the minimum MCM model $\mathcal{G}$ for $\mathcal{U}$.
Following training of the generative model, the model may be used to generate predictions in the observation space via draws from the latent space.
Since our representation of the latent space was constructed according to the minimum MCM graph $\mathcal{G}$, the resulting predictions can be viewed as causally informed; i.e., they are observations generated from the estimated distribution of the latent primary causes of the measurement variables.

\subsection{Identifiability of minimum MCM graphs and ECC-model equivalence}
\label{sec:identifiability}

While the UDG is identifiable, there may exist multiple minimum MCM graphs that yield the same UDG.
This is because an undirected graph may have multiple, distinct minimum edge clique covers (see, for instance, \Cref{ex:nonident} and surrounding discussion in \Cref{appsec: identifiability}).
In other words, similar to DAGs, minimum MCM graphs may be equivalent when provided with only observational data.

\begin{definition}
	\label{def: ECC obs equiv}
	We say that two minimum MCM graphs $\mathcal{G} = \langle [n] \cup L, E\rangle$ and $\mathcal{G}^\prime = \langle [n]\cup L^\prime, E^\prime\rangle$ are \emph{ECC-observationally equivalent} if $i$ and $j$ are $d$-separated given $\emptyset$ in $\mathcal{G}$ if and only if they are $d$-separated given $\emptyset$ in $\mathcal{G}^\prime$.
\end{definition}

While there exist equivalence classes of minimum MCM graphs containing multiple elements, there also exist classes that are singletons; in other words, there exist undirected graphs (UDGs) with a unique minimum edge clique cover.
For such UDGs, the minimum MCM graph is identifiable.

\begin{theorem}
	\label{thm: ident}
	Suppose that the data-generating distribution is Markov to a minimum MCM graph $\mathcal{G}$, and let $\mathcal{U}$ be the UDG induced by $\mathcal{G}$.
	Then the DAG $\mathcal{G}$ is identifiable from the data-generating distribution if:
	\begin{enumerate}
		\item $\mathcal{U}$ admits a unique minimum edge clique cover, and
		\item $\mathcal{G}$ is \emph{measurement-faithful} to the data-generating distribution \citep[Def.~5]{markham20mcm}.
	\end{enumerate}
\end{theorem}

Proofs of this theorem and the following corollary are given in \Cref{appsec: identifiability}.
A concrete example of a class of practically relevant models that admit a unique minimum edge clique cover are those that satisfy the 1-pure-child assumption: For each latent \(l_i\) in \(\mathcal{G}\) there exists a measurement \(i^*\) such that \(\mathrm{pa}_{\mathcal{G}}(i^*) = \{l_i\}\).
This assumption is widely used in machine learning \citep[e.g., as in][]{donoho2003does,arora2012learning,bing2020adaptive,moran2023identifiable}, and also appears in biological applications of factor models \citep{bing2019latent,bing2022essential,rahimikollu2024slide}.
This is also the main condition assumed by \citet{kim23structure}, who study linear factor models.

\begin{corollary}
	\label{cor:ident}
	If the data-generating distribution is Markov to a minimum MCM graph $\mathcal{G}$ satisfying the 1-pure-child assumption, and $\mathcal{G}$ is measurement-faithful to the data-generating distribution \citep[Def.~5]{markham20mcm}, then \(\mathcal{G}\) is identifiable.
\end{corollary}

Of course, this is just one example, and \Cref{thm: ident} shows that these are not the only models to which the identifiability result applies.
An explicit construction of a model that admits a unique minimum edge clique cover but does not satisfy the 1-pure-child assumption is given in \Cref{ex:impure} (\Cref{appsec: identifiability}).
We emphasize once again that these results impose no functional restrictions, and in particular apply to general nonlinear models without additive noise or auxiliary information, including potentially mixed discrete variables.

\begin{remark}
	\label[remark]{rem:latent-ident}
	Intuitively, a minimum MCM graph may be viewed as a summary of the confounding ancestors, and in particular the ``causal sources'', of a set of measurement variables.
	Imagine an arbitrary DAG, and consider its sink nodes to be the measurement variables---this is a (non-minimum) MCM graph.
	For each clique in a minimum edge clique cover of the corresponding UDG, the collection of common confounding ancestors that are relevant to that clique in the MCM graph get projected into a single latent factor in the minimum MCM graph.
	In this sense, the identified latent variables can be interpreted as a minimal projection of the underlying confounding structure onto the observational level.

	At the distributional level, one could consider the pushforward measure induced by this projection.
	For example, in the linear Gaussian case, this would look like principal component analysis, and one could make use of linearity and Gaussianity to characterize identifiability of the minimal latent distribution;
	in the general NCFA setting, the corresponding latent representation may be nonlinear and nonparametric, but it still plays the same conceptual role as a compressed summary of the common causes.

	Moreover, once the minimum MCM graph is identified, existing identifiability results \citep{kivva2022identifiability,moran2023identifiable,wang2021posterior} for deep generative models imply that, under suitable assumptions on the decoder and latent prior, the associated latent block distributions are identifiable up to the usual reparameterization ambiguities.
	Thus, graph identifiability provides the structural part of the argument, while the generative model identifiability theory supplies the distributional part.
\end{remark}

\subsection{Learning}
\label{sec:algorithm}

The data to UDG to NCFA-graph and model pipeline of the previous subsections can be formalized into Neuro-Causal Factor Analysis (NCFA) as \Cref{alg:ncfa}:

\setlength{\algomargin}{1.5em}
\begin{algorithm2e}
	\caption{Neuro-Causal Factor Analysis (NCFA)}
	\label{alg:ncfa}
	\DontPrintSemicolon
	\LinesNumbered
	\KwIn{sample \(S\) of measurement variables \(M\)}
	\Parameter{significance level \(\alpha\), latent degrees of freedom \(\lambda\)}
	\KwOut{neuro-causal factor model \(\langle \widetilde{\mathcal{G}}, f_{[n]} \rangle\), with NCFA graph \(\widetilde{\mathcal{G}}\) and loading functions \(f_{[n]}\)}
	Obtain undirected dependence graph \(\mathcal{U}\) via pairwise marginal independence tests with threshold \(\alpha\)\;\label{line:U}

	Identify minimum edge clique cover \(\mathcal{C}\) of $\mathcal{U}$ and construct corresponding minimum MCM graph $\mathcal{G}$\; \label{line: min clique cov}

	Assign remaining $\lambda - \lvert \mathcal{C}\rvert$ latents to the cliques in $\mathcal{C}$ to produce the NCFA-graph $\widetilde{\mathcal{G}}$\label{line:lambda}\;

	Estimate functions \(f_{[n]}\) using a VAE constrained by \(\widetilde{\mathcal{G}}\)\label{line:L-func}\;
\end{algorithm2e}

To estimate the UDG, pairwise marginal independence tests are used.
Starting with the complete graph, the edge $i \edge j$ is removed whenever $M_i$ and $M_j$ are deemed independent, i.e., according to a test with statistics such as distance covariance \citep{szekely2007measuring,markham2022distance} or xi correlation \citep{chatterjee2021new,10.1093/biomet/asac048}.
Since these independence tests are estimated from finite data, they are subject to Type~I and Type~II errors, so the estimated UDG $\widehat{\mathcal{U}}$ may contain spurious edges or omit true ones;
thus, although \Cref{thm: ident} establishes identifiability when the marginal independence relations are known, the graph recovered from a finite sample need not coincide with the true minimum MCM graph.
A minimum edge clique cover is then identified for the estimated UDG $\widehat{\mathcal{U}}$.
In general, this is an NP-hard problem, however there are both exact algorithms that work well for small graphs and heuristic algorithms that scale to large graphs \citep{gramm2009,conte2020,ullah2022computing}.

Once a minimum edge clique cover is identified, the corresponding NCFA graph with $\lambda$ latent degrees of freedom is constructed.
Here, we ensure that at every clique in the minimum edge clique cover of $\widehat{\mathcal{U}}$ is assigned at least one latent variable.
The remaining $\lambda - k$ latents are then distributed uniformly over the cliques.
Finally, a VAE for the functional relations specified by the NCFA-graph is trained.
One could, in principle, alternatively use any deep generative model.

Since NCFA constructs its model via the MCM graph $\widehat{\mathcal{U}}$, the estimated factors (i.e., joint distributions) $f(L_i)$ in the factorization of the latent distribution represent the distributions for the primary causes of the measurement variables to which the latent nodes in $\mathcal{L}_i$ are connected.
This yields a factor analysis model in which the latent factors can justifiably be causally interpreted.
Furthermore, each causal latent $L_i = (Z_{i,1},\ldots,Z_{i,k_i})$ is itself Gaussian, $L_i \sim \mathcal{N}(0, I_{k_i})$, for any choice of $k_i$.
Increasing $k_i$ grows the dimension of this Gaussian input to the nonlinear decoder $f_i$, whose pushforward can represent arbitrarily non-Gaussian contributions to $M_i$.
Hence, the estimated factors have a causal interpretation while additionally being as nonlinear as necessary.

\section{Experiments}
\label{sec:appl}

\begin{table}[t]
	\centering
	\resizebox{\textwidth}{!}{%
		\insertsimulated%
	}
	\caption{Results on linear Gaussian simulated data.}
	\label{tab:sim-results}
\end{table}

We now evaluate NCFA on synthetic and real data, comparing against classical factor analysis (FA), a fully connected variational autoencoder (VAE), and a linear Gaussian minimum MCM model (LG-minMCM).
Thus, we compare both sparse causal structure versus fully connected models (NCFA, LG-minMCM versus VAE, FA) as well as linear Gaussian vs nonparametric (FA, LG-minMCM versus VAE, NCFA).
For quantitative evaluation we report, for each method:
\begin{itemize}
	\item \textbf{MSE}: mean squared reconstruction error on held-out data;
	\item \textbf{MCC}: mean correlation coefficient between matched pairs of learned and ground truth latents (matched to handle sign/permutation indeterminacy; see our implementation for details), a standard identifiability metric in the ICA and causal representation learning literature \citep{hyvarinen2017nonlinear,khemakhem2020variational,gamella2025sanity};
	\item \textbf{dCor}: distance correlation between joint learned and ground truth latent distributions---unlike MCC's linear matching, dCor also captures nonlinear reparameterizations, relevant given NCFA's nonlinear decoder;
	\item \textbf{SFD}: the shared factor distance between the learned and ground truth minimum MCM graphs (\Cref{sec:evaluation-metrics}), quantifying how well the causal structure is learned;
	\item \textbf{Time}: average wall-clock runtime; and
	\item \textbf{\# Params}: the number of trainable parameters in the generative model.
\end{itemize}
Median values of all metrics are reported, plus or minus the median absolute deviation, computed over 101 random seeds.
The first four metrics are computed on held out data, using a 70/30 train/test split.

LG-minMCM and NCFA both use the 1-pure-child assumption, ensuring fast recovery of the minimum edge clique cover.
LG-minMCM estimates the BIC-optimal UDG, while NCFA uses xi correlation \citep{chatterjee2021new,10.1093/biomet/asac048} for independence tests.

All four methods compute held-out MSE identically (see our implementation for details);
FA here is the saturated, fully connected linear Gaussian MCM ($K=n$), the dense counterpart to LG-minMCM's sparse structure, and its large MSE reflects this model's unidentifiability (\Cref{sec:factor-analysis}).

For the synthetic experiments, we generate minimum MCM graphs over $10$ measurement variables and \(3\) latents, with edge probability \(0.3\).
On top of each such graph, we generate a dataset containing \(10\,000\) observations from a linear Gaussian factor model.
\Cref{tab:sim-results} shows the results.
The assumptions for LG-minMCM are all satisfied, so it performs best on latent recovery (MCC and dCor), while being faster and using fewer parameters than the other methods.
Nonetheless, the VAE is able to achieve better reconstruction (MSE), at the cost of worse latent recovery.
This gap is not attributable to errors in graph estimation, since LG-minMCM's assumptions are already satisfied here;
rather, it reflects a capacity difference: the fully connected VAE has roughly $200\times$ as many parameters as LG-minMCM (7350 vs.\ 35) and no sparsity constraint, letting it fit the reconstruction task more flexibly at the cost of its latents no longer corresponding to the true causal factors.
NCFA attains comparable latent recovery to LG-minMCM and slightly better reconstruction, with run time and number of parameters comparable to the VAE.
Thus, NCFA performs reasonably here, almost as well as the simple (correctly specified) parametric model and better than the VAE, which isn't designed for latent recovery.

Here, we treat sink nodes as measurement variables and the rest as latents, resulting in the ground truth minimum MCM graph in \Cref{fig:lt-b}.
The data is non-Gaussian, so the LG-minMCM is misspecified.
\Cref{tab:real-results} summarizes the results.
Here, NCFA attains reconstruction performance comparable to VAE and the best latent recovery among all methods (in terms of MCC, though its dCor is nearly tied with VAE's, and LG-minMCM attains better structural recovery (SFD).
The high MCC and low SFD are empirical corroboration of the identifiability results in \Cref{thm: ident} and \Cref{rem:latent-ident}.

\begin{figure}[t]
	\centering
	% ---------- (a) Ground-truth DAG ---------- 
\subfigure[Ground-truth DAG.]{%
	\label{fig:lt-a}%
	\resizebox{0.32\textwidth}{!}{%
		\begin{tikzpicture}[>=stealth, auto=false, node distance=1.2cm]
			\tikzstyle{lt}=[circle, inner sep=0.06em]
			% Layer 0 (sources, top)
			\node[lt] (R)  at (0,   3.7) {$R$};
			\node[lt] (G)  at (0.6, 3.7) {$G$};
			\node[lt] (B)  at (1.2, 3.7) {$B$};
			\node[lt] (T1) at (-1.6,3.7) {$\theta_1$};
			\node[lt] (T2) at (-0.8,3.7) {$\theta_2$};
			% Lens outputs (top right)
			\node[lt] (L11) at (2.4,3.7) {$L_{11}$};
			\node[lt] (L12) at (3.0,3.7) {$L_{12}$};
			\node[lt] (L21) at (3.6,3.7) {$L_{21}$};
			\node[lt] (L22) at (4.2,3.7) {$L_{22}$};
			\node[lt] (L31) at (4.8,3.7) {$L_{31}$};
			\node[lt] (L32) at (5.4,3.7) {$L_{32}$};
			% Layer 1 (near-field, middle)
			\node[lt] (C)  at (0.3,1.2) {$\tilde{C}$};
			\node[lt] (I1) at (0.9,1.2) {$\tilde{I}_1$};
			\node[lt] (I2) at (1.5,1.2) {$\tilde{I}_2$};
			\node[lt] (V1) at (2.1,1.2) {$\tilde{V}_1$};
			\node[lt] (V2) at (2.7,1.2) {$\tilde{V}_2$};
			% Layer 2 (far-field + polarizer states, bottom)
			\node[lt] (I3)  at (1.2,-0.6) {$\tilde{I}_3$};
			\node[lt] (V3)  at (0.6,-0.6) {$\tilde{V}_3$};
			\node[lt] (T1t) at (-1.8,-0.6) {$\tilde{\theta}_1$};
			\node[lt] (T2t) at (-1.2,-0.6) {$\tilde{\theta}_2$};
			% Helper control points for RGB -> (V3,I3)
			\coordinate (auxR1) at (-0.5,1.6);
			\coordinate (auxR2) at (-0.4,1.4);
			\coordinate (auxG1) at (-0.45,1.6);
			\coordinate (auxG2) at (-0.35,1.4);
			\coordinate (auxB1) at (-0.40,1.6);
			\coordinate (auxB2) at (-0.30,1.4);
			% RGB -> C, I1, I2, V1, V2
			\foreach \src in {R,G,B} {
					\draw[->] (\src) -- (C);
					\draw[->] (\src) -- (I1);
					\draw[->] (\src) -- (I2);
					\draw[->] (\src) -- (V1);
					\draw[->] (\src) -- (V2);
				}
			% RGB -> V3, I3 (curved, left of C)
			\draw[->] (R) .. controls (auxR1) .. (V3);
			\draw[->] (R) .. controls (auxR2) .. (I3);
			\draw[->] (G) .. controls (auxG1) .. (V3);
			\draw[->] (G) .. controls (auxG2) .. (I3);
			\draw[->] (B) .. controls (auxB1) .. (V3);
			\draw[->] (B) .. controls (auxB2) .. (I3);
			% theta nodes
			\draw[->] (T1) -- (T1t);
			\draw[->] (T2) -- (T2t);
			\draw[->] (T1) to[out=-90,in=160] (V3);
			\draw[->] (T1) to[out=-90,in=160] (I3);
			\draw[->] (T2) to[out=-90,in=170] (V3);
			\draw[->] (T2) to[out=-90,in=170] (I3);
			% Near-field -> far-field
			\draw[->] (V1) -- (I3);
			\draw[->] (V1) -- (V3);
			\draw[->] (V2) -- (I3);
			\draw[->] (V2) -- (V3);
			% Lens outputs -> intensities
			\draw[->] (L11) -- (I1);
			\draw[->] (L12) -- (I1);
			\draw[->] (L21) -- (I2);
			\draw[->] (L22) -- (I2);
			\draw[->] (L31) to[out=-120,in=30] (I3);
			\draw[->] (L32) to[out=-120,in=30] (I3);
		\end{tikzpicture}%
	}%
}\quad
% ---------- (b) Coarsened DAG (grouped) ----------
\subfigure[Ground-truth minMCM.]{%
	\label{fig:lt-b}%
	\resizebox{0.32\textwidth}{!}{%
		\begin{tikzpicture}[>=stealth, auto=false, node distance=1.2cm]
			\tikzstyle{lt}=[circle, inner sep=0.06em]
			% Sources
			\node[lt] (T1) at (-1.0, 2.0) {$\theta_1$};
			\node[lt] (T2) at (0.0, 2.0) {$\theta_2$};
			\node[lt] (S)  at (1.2, 2.0) {$S$};
			% Sinks
			\node[lt] (T1t) at (-1.8, 0.0) {$\tilde{\theta}_1$};
			\node[lt] (T2t) at (-1.0, 0.0) {$\tilde{\theta}_2$};
			\node[lt] (C)  at (0.0, 0.0) {$\tilde{C}$};
			\node[lt] (I1)  at (1.0, 0.0) {$\tilde{I}_1$};
			\node[lt] (I2)  at (1.8, 0.0) {$\tilde{I}_2$};
			\node[lt] (I3)  at (2.6, 0.0) {$\tilde{I}_3$};
			\node[lt] (V3)  at (3.6, 0.0) {$\tilde{V}_3$};
			% Edges
			\draw[->] (T1) -- (T1t);
			\draw[->] (T1) -- (I3);
			\draw[->] (T1) -- (V3);
			\draw[->] (T2) -- (T2t);
			\draw[->] (T2) -- (I3);
			\draw[->] (T2) -- (V3);
			\draw[->] (S) -- (C);
			\draw[->] (S) -- (I1);
			\draw[->] (S) -- (I2);
			\draw[->] (S) -- (I3);
			\draw[->] (S) -- (V3);
		\end{tikzpicture}%
	}%
}
% ---------- (c) Coarsened DAG (ungrouped) ----------
\subfigure[Learned minMCM.]{%
	\label{fig:lt-c}%
	\resizebox{0.32\textwidth}{!}{%
		\begin{tikzpicture}[>=stealth, auto=false, node distance=1.2cm]
			\tikzstyle{lt}=[circle, inner sep=0.06em]
			% Sources
			\node[lt] (T1) at (-1.0, 2.0) {$\theta_1$};
			\node[lt] (T2) at (0.0, 2.0) {$\theta_2$};
			\node[lt] (S)  at (1.2, 2.0) {$S$};
			% Sinks
			\node[lt] (T1t) at (-1.8, 0.0) {$\tilde{\theta}_1$};
			\node[lt] (T2t) at (-1.0, 0.0) {$\tilde{\theta}_2$};
			\node[lt] (C)  at (0.0, 0.0) {$\tilde{C}$};
			\node[lt] (I1)  at (1.0, 0.0) {$\tilde{I}_1$};
			\node[lt] (I2)  at (1.8, 0.0) {$\tilde{I}_2$};
			\node[lt] (I3)  at (2.6, 0.0) {$\tilde{I}_3$};
			\node[lt] (V3)  at (3.6, 0.0) {$\tilde{V}_3$};
			% Edges (false negative in (dashed, gray))
			\draw[->] (T1) -- (T1t);
			\draw[->, dashed, gray] (T1) -- (I3);
			\draw[->, dashed, gray] (T1) -- (V3);
			\draw[->] (T2) -- (T2t);
			\draw[->] (T2) -- (I3);
			\draw[->, dashed, gray] (T2) -- (V3);
			\draw[->] (S) -- (C);
			\draw[->] (S) -- (I1);
			\draw[->] (S) -- (I2);
			\draw[->] (S) -- (I3);
			\draw[->] (S) -- (V3);
		\end{tikzpicture}%
	}%
}
%%% Local Variables:
%%% mode: LaTeX
%%% TeX-master: "main"
%%% End:
	\caption{Panels show (a) the full ground truth DAG, (b) the true minMCM, and (c) an example estimated minMCM.
		False negatives in \textcolor{gray}{dashed gray}; no false positives.}
	\label{fig:lt}
\end{figure}

\begin{table}
	\centering
	\resizebox{\textwidth}{!}{%
		\insertcausalchamber%
	}
	\caption{Results on real data from the causal chambers.}
	\label{tab:real-results}
\end{table}

\section{Conclusion}
\label{sec:conclusion}

In summary, NCFA gives a graphical route to factor identifiability.
Its main limitations are the interpretability of the unique minimum edge clique cover assumption and the extent to which the learned factors are genuinely causal.
Future work includes exploring other graphical identifiability results for factor analysis \citep{drton2025algebraic,sturma2025matching,lee2025identifiability,gu2025blessing}, replacing the constraint-based step with sparsity regularization for end-to-end training \citep{moran2023identifiable,zhang2026discrete}, and using interventions to improve interpretability, identifiability, and out-of-distribution generalization \citep{jiang2023learning,markham2026}.

\acks{
	The work by AM was partially supported by Novo Nordisk Foundation grant number NNF20OC0062897 and also by the Pioneer Centre for Statistical and computational Methods for Advanced Research to Transform Biomedicine (SMARTbiomed), DNRF grant number P4.
	Part of this research was performed while AM, BA, and LS were visiting the Institute for Mathematical and Statistical Innovation (IMSI), which is supported by the National Science Foundation (Grant No.~DMS-1929348), and the authors thank the organizers of the IMSI long program \textit{Algebraic Statistics and Our Changing World}.
	This research was supported in part through the computational resources and staff contributions provided for the Mercury high performance computing cluster at The University of Chicago Booth School of Business which is supported by the Office of the Dean.
}

\bibliography{main}

\newpage
\appendix
\counterwithin{figure}{section}
\counterwithin{table}{section}

\onecolumn

\section{MeDIL Causal Models}
\label[appendix]{appsec: medil models}
The NCFA models used in this paper are a subfamily of the models known as measurement dependence inducing latent (MeDIL) causal models, originally introduced in \citep{markham20mcm}.
For contextualization purposes, we give a brief description of MeDIL causal models here.

A \emph{MCM graph} is a triple $\mathcal{G} = \langle M, L, E\rangle$ where $M = \{1,\ldots, n\}$ and $L = \{l_1,\ldots, l_{K}\}$ are disjoint sets of vertices corresponding, respectively, to observed variables and latent variables, and $E$ is the collection of directed edges between the nodes.
In a MCM graph we require that the nodes in $M$ are all \emph{sinks} (i.e., have out-degree $0$) and that the edges $E$ are such that $\mathcal{G}$ is a \emph{directed acyclic graph} (DAG).
A distribution belongs to the \emph{Measurement Dependence Inducing Latent (MeDIL) Causal Model} if it factorizes according to $\mathcal{G}$, i.e., if the probability density function (or probability mass function) satisfies \[ f(x_1,\ldots, x_{n}, x_{l_1},\ldots, x_{l_{K}}) = \prod_{i = 1}^{n}f(x_i | x_{\textrm{pa}_{\mathcal{G}}(i)})\prod_{j = 1}^{K}f(x_{l_j} | x_{\textrm{pa}_{\mathcal{G}}(l_i)}).
\]
Since we do not observe the latent variables, but only the observed variables, $X_1,\ldots, X_{n}$, the distribution of interest is the marginal obtained by integrating out the latents:
\begin{equation}
	\label{eqn: medil factorization}
	f(x_1,\ldots, x_{n}) = \int_{\mathcal{X}_{L}}\prod_{i = 1}^{n}f(x_i | x_{\textrm{pa}_{\mathcal{G}}(i)})\prod_{j = 1}^{K}f(x_{l_j} | x_{\textrm{pa}_{\mathcal{G}}(l_j)})dx_L.
\end{equation}
The MeDIL model $\mathcal{M}(\mathcal{G})$ consists of all observable distributions that arise as this marginal.

In general, MCM graphs may have a complex directed acyclic structure over the latents that induce correlations amongst the observables.
However, since we only consider the observed variables, we may reduce to a minimum MCM representation of the distribution.
In a \emph{minimum MCM graph}, we further assume that the latent variables are \emph{source nodes} (i.e., have in-degree $0$) and out-degree at least $1$.
Hence, a \emph{minimum MeDIL model}---i.e., a MeDIL model that factorizes as in equation~\eqref{eqn: medil factorization}---explains the associations among the observed variables in the simplest possible way: via a collection of independent latents.
These independent latents can be thought of as source nodes in the original MCM graph $G$, and the associated minimum MCM graph as the graph produced after we marginalize out all non-source latent variables.
While learning the complex causal structure on the latents in a general MCM graph may be intractable, one can apply existing methods to estimate the simpler minimum MCM structure.
Intuitively, one can think of learning the minimum MCM graph of a model as identification of the "primary causes" of the associations between the observed variables.

It follows from the factorization of a distribution according to a minimum MCM graph that the only conditional independence constraints amongst the observed variables are marginal independence constraints.
These, in turn, encode the condition that two nodes do not have a shared (latent) parent.
From the causal perspective, we see that two nodes in the system are marginally independent if and only if they share a latent common cause.
Hence, minimum MeDIL models are the natural representation of Reichenbach's Common Cause Principle.

It follows that a natural representation of the model using only the observed variables is via an undirected graph $\mathcal{U} = \langle M, E\rangle$, where $i - j\notin E$ if and only if $i$ and $j$ are independent in the joint distribution of the observed variables.
We call this graph the \emph{unconditional dependence graph} (UDG) of the model.
A \emph{minimum edge clique cover} of the undirected graph $U$ is a collection of cliques $\mathcal{C} = \{C_1,\ldots, C_K\}$ for which every pair $i,j$ satisfying $i - j\in E$ is contained in at least one clique in $\mathcal{C}$ and there exists no set of cliques with this property that has cardinality smaller than $|\mathcal{C}|$.
Since a UDG depends only on the observed variables, it is possible to learn a UDG from the available data via pairwise marginal independence tests.
A minimum MCM graph that captures the marginal independence structure encoded via the UDG is then the minimum MCM graph where we have one latent $l_i$ for each clique $C_i \in \mathcal{C}$ and $l_i$ has a directed arrow to each node $j\in C_i$ (and no other adjacencies).

To learn the assignment of latents producing a minimum MCM graph from a UDG $\mathcal{U}$, we must learn a minimum edge clique cover of $\mathcal{U}$.
The problem of identifying a minimum edge clique cover of an undirected graph is NP-hard.
There are exact algorithms \citep{gramm2009,ullah2022computing} showing it to be NP-complete and fixed-parameter tractable.
However, there are also polynomial-time heuristic algorithms \citep{conte2020} that can efficiently handle very large graphs.

\section{Identifiability of minimum MCM graphs: Examples and Counterexamples}
\label[appendix]{appsec: identifiability}

\begin{proof}[Proof of Theorem~\ref{thm: ident}]
	Because the data-generating distribution is Markov to \(\mathcal{G}\), the use of any consistent nonparametric test for independence (e.g., \citep{chatterjee2021new}) is guaranteed to identify all marginal independence statements in the limit.
	By measurement-faithfulness (Condition 2), these coincide exactly with the edges of \(\mathcal{U}\), the UDG induced by \(\mathcal{G}\).
	Using \citep[Proposition 7]{markham20mcm} to associate \(\mathcal{U}\) with a minimum edge clique cover recovering \(\mathcal{G}\), the uniqueness of this minimum edge clique cover (Condition 1) guarantees the identifiability of \(\mathcal{G}\).
\end{proof}

\begin{proof}[Proof of Corollary~\ref{cor:ident}]
	We show the 1-pure-child assumption gives Condition 1 of Theorem~\ref{thm: ident};
	Condition 2 is assumed directly in the corollary's hypothesis.
	The association of a minimum MCM graph \(\mathcal{G}\) with a minimum edge clique cover of \(\mathcal{U}\) \citep[Proposition 7]{markham20mcm} induces an association between a maximum independent set in \(\mathcal{U}\) and a set of pure children in \(\mathcal{G}\) with different parents.
	Observe that \(\mathcal{G}\) satisfies the 1-pure-child assumption if and only if there exists a maximum independent set \(\{i^*\}_{i=1}^K\) in \(\mathcal{U}\) consisting of one pure child for each latent.
	In other words, \(\mathcal{G}\) satisfies the 1-pure-child assumption if and only if the independence number (i.e., size of a maximum independent set, which is the number latents with at least one pure child) and intersection number (i.e., the number of cliques in a minimum edge clique cover, which is the number of latents) of its corresponding \(\mathcal{U}\) are equal.
	It follows from Lemma~2.3.5 of~\citet{deligeorgaki2022combinatorial} that \(\mathcal{U}\) has a unique minimum edge clique cover, giving Condition 1.
	Hence \(\mathcal{G}\) is identifiable by Theorem~\ref{thm: ident}.
\end{proof}

The following examples illustrate the types of models that can be identified via Theorem~\ref{thm: ident}.

\begin{example}
	\label{ex:pure}
	Graphs admitting a \emph{pure measurement variable} in each clique of a minimum edge clique cover will have a unique minimum edge clique cover, making the resulting minimum MCM graph identifiable (see Theorem~\ref{thm: ident}).
	From a graphical perspective, we call this condition the \emph{1-pure child constraint} since it insists that each latent node $l_i$ in the minimum MCM graph has (at least) one child $i^*$ which has no other parents than $l_i$ (i.e., $l_i \rightarrow i^*$ is an edge of $\mathcal{G}$ and $i^*$ has no other adjacencies in $\mathcal{G}$).
\end{example}

\begin{example}
	\label{ex:impure}
	This example shows that the 1-pure-child constraint in Example~\ref{ex:pure} is not necessary for a UDG to have a unique minimum edge clique cover, and thus Theorem~\ref{thm: ident} ensures identifiability of models violating this constraint.
	In particular, the following UDG $\mathcal{U}$ has the unique minimum edge clique cover $\mathcal{C} = \{\{1,4\},\{2,5\},\{3,6\},\{4,5,6\}\}$ and hence the corresponding minimum MCM graph $\mathcal{G}$ is identifiable:
	\begin{center}
		\begin{tikzpicture}[scale=.4] \node at (-2,-4) {$\mathcal{U}$};
			\begin{scope}[execute at begin node=$, execute at end node=$] \Vertex{1} \SO(1){4} \EA(4){5} \SO(5){6} \NO(5){2} \SO(6){3} \Edges(1, 4) \Edges(2, 5) \Edges(3, 6) \Edges(4, 5) \Edges(5, 6) \Edges(6, 4)
			\end{scope}
		\end{tikzpicture}
		\hspace{1.5cm} \tikzstyle{EdgeStyle} = [->,>=stealth']
		\begin{tikzpicture}[scale=.5] \node at (-2,1) {$\mathcal{G}$};
			\begin{scope}[execute at begin node=$, execute at end node=$] \Vertex{1} \EA(1){2} \EA(2){3} \EA(3){4} \EA(4){5} \EA(5){6} \NO(2){l_1} \EA(l_1){l_2} \EA(l_2){l_3} \EA(l_3){l_4} \Edges(l_1, 1) \Edges(l_1, 4) \Edges(l_2, 2) \Edges(l_2, 5) \Edges(l_3, 3) \Edges(l_3, 6) \Edges(l_4, 4) \Edges(l_4, 5) \Edges(l_4, 6)
			\end{scope}
		\end{tikzpicture}
	\end{center}
	Note here that the latent variables $l_1,l_2$ and $l_3$ each have a pure child ($1,2$ and $3$, respectively).
	However, the latent $l_4$ does not.
	Hence $\mathcal{G}$ is identifiable but does not satisfy the 1-pure-child constraint.
\end{example}

\begin{example}
	\label{ex:discrete}
	Since previous work mainly focuses on continuous data, we point out that our assumptions are easily satisfied by models with discrete variables.
	For example, Assumption~2 in Theorem~\ref{thm: ident} is known to hold generically \citep{meek1995strong} for discrete (categorical) DAGs.
	Specifically, \citet{meek1995strong} showed the existence of discrete distributions that are faithful to any given DAG for any choice of state spaces of the discrete variables.
	These same distributions necessarily fulfill Assumption~2 in Theorem~\ref{thm: ident}.
	Moreover, any marginal independence relation $X_k \independent X_\ell$ that is not encoded by the DAG is satisfied by the distributions represented as points in the open probability simplex that vanish on a collection of $2\times 2$ minors of the matrix of parameters encoding the joint probabilities of $X_k$ and $X_\ell$ \citep{garcia2005algebraic}.
	Using computational algebra software as described in \citep{garcia2005algebraic}, one can check for small examples of DAG models that the intersection of this set of points satisfying the polynomial constraints corresponding to $X_k \independent X_\ell$ with the points vanishing on the polynomials corresponding to the CI relations encoded by the DAG has a lower dimension than the DAG model.
	This implies that almost all points in the DAG model, including points that are not faithful to the DAG, still satisfy Assumption~2 in Theorem~\ref{thm: ident}.
	Hence, we may observe that Assumption~2 is strictly weaker than faithfulness.
	This, combined with the observations in Examples~\ref{ex:pure}-\ref{ex:impure} provides a broad family of discrete models that satisfy our identifiability conditions, and that cannot be represented by existing work (see Appendix~\ref{app:detailedcomp}).
\end{example}

\begin{example}
	\label{ex:nonident}
	An example of a minimum MCM graph that cannot be identified because it admits multiple edge clique covers is the following:
	\begin{center}
		\tikzstyle{EdgeStyle} = [->,>=stealth']
		\begin{tikzpicture}[scale=.5] \node at (-2,1) {$\mathcal{G}$};
			\begin{scope}[execute at begin node=$, execute at end node=$] \Vertex{1} \EA(1){2} \EA(2){3} \EA(3){4} \EA(4){5} \EA(5){6} \NO(2){l_1} \EA(l_1){l_2} \EA(l_2){l_3} \EA(l_3){l_4} \Edges(l_1, 1) \Edges(l_1, 3) \Edges(l_1, 4) \Edges(l_2, 1) \Edges(l_2, 5) \Edges(l_2, 6) \Edges(l_3, 2) \Edges(l_3, 3) \Edges(l_3, 6) \Edges(l_4, 2) \Edges(l_4, 4) \Edges(l_4, 5)
			\end{scope}
		\end{tikzpicture}
	\end{center}
	\sloppy
	The UDG of $\mathcal{G}$ is the edge graph $\mathcal{U}$ of the octahedron which admits exactly two minimum edge clique covers: $\mathcal{C}_1 = \{\{1,3,4\},\{1,5,6\},\{2,3,6\},\{2,4,5\}\}$ and $\mathcal{C}_2 = \{\{1,3,6\},\{1,4,5\},\{2,3,4\},\{2,5,6\}\}$.
	$\mathcal{C}_1$ recovers the minimum MCM graph $\mathcal{G}$, whereas $\mathcal{C}_2$ produces the minimum MCM graph $\mathcal{H}$, depicted below:
	\begin{center}
		\tikzstyle{EdgeStyle} = [->,>=stealth']
		\begin{tikzpicture}[scale=.5]
			\node at (-2,1) {$\mathcal{H}$};
			\begin{scope}[execute at begin node=$, execute at end node=$]
				\Vertex{1} \EA(1){2} \EA(2){3} \EA(3){4} \EA(4){5} \EA(5){6}
				\NO(2){l_1} \EA(l_1){l_2} \EA(l_2){l_3} \EA(l_3){l_4}
				\Edges(l_1, 1) \Edges(l_1, 3) \Edges(l_1, 6)
				\Edges(l_2, 1) \Edges(l_2, 4) \Edges(l_2, 5)
				\Edges(l_3, 2) \Edges(l_3, 3) \Edges(l_3, 4)
				\Edges(l_4, 2) \Edges(l_4, 5) \Edges(l_4, 6)
			\end{scope}
		\end{tikzpicture}
	\end{center}
	This issue arises due to the spherical structure of the UDG $\mathcal{U}$, depicted on the left below.
	In the remaining two figures, the each shaded triangle represents the children of a latent node (i.e., a clique in a minimum edge clique cover) of a possible MCM graph with associated UDG $\mathcal{U}$.
	The middle figure represents the minimum MCM graph $\mathcal{G}$ and the rightmost figure represents the alternative $\mathcal{H}$.
	\begin{center}
		\begin{tikzpicture}[thick,scale=0.7]

			%---NODES---
			\node (1) at (0,2,0)  {$1$};
			\node (2) at (-2,0,-2) {$6$};
			\node (3) at (1,0,-1)  {$3$};
			\node (4) at (2,0,2) {$4$};
			\node (5) at (-1,0,1) {$5$};
			\node (6) at (0,-2,0) {$2$};

			%---EDGES---
			\draw[-]   (1) -- (2) ;
			\draw[-]   (1) -- (3) ;
			\draw[-]   (1) -- (4) ;
			\draw[-]   (1) -- (5) ;
			\draw[-]   (6) -- (2) ;
			\draw[-]   (6) -- (3) ;
			\draw[-]   (6) -- (4) ;
			\draw[-]   (6) -- (5) ;
			\draw[-]   (2) -- (3) ;
			\draw[-]   (3) -- (4) ;
			\draw[-]   (4) -- (5) ;
			\draw[-]   (5) -- (2) ;

		\end{tikzpicture}
		\hspace{1cm}
		\begin{tikzpicture}[thick,scale=0.7]

			\draw[fill=blue!30] (0,-2,0) -- (-2,0,-2) -- (1,0,-1) -- (0,-2,0) -- cycle  ;
			\draw[fill=blue!30] (0,-2,0) -- (2,0,2) -- (-1,0,1) -- (0,-2,0) -- cycle  ;

			\draw[fill=blue!30] (0,2,0) -- (1,0,-1) -- (2,0,2) -- (0,2,0) -- cycle  ;
			\draw[fill=blue!30] (0,2,0) -- (-2,0,-2) -- (-1,0,1) -- (0,2,0) -- cycle  ;

			%---NODES---
			\node (1) at (0,2.3,0)  {$1$};
			\node (2) at (-2.3,0,-2) {$6$};
			\node (3) at (1.3,0,-1)  {$3$};
			\node (4) at (2.3,0,2) {$4$};
			\node (5) at (-1.3,0,1) {$5$};
			\node (6) at (0,-2.3,0) {$2$};

		\end{tikzpicture}
		\hspace{1cm}
		\begin{tikzpicture}[thick,scale=0.7]

			\draw[fill=blue!30] (0,-2,0) -- (1,0,-1) -- (2,0,2) -- (0,-2,0) -- cycle  ;
			\draw[fill=blue!30] (0,-2,0) -- (-2,0,-2) -- (-1,0,1) -- (0,-2,0) -- cycle  ;

			\draw[fill=blue!30] (0,2,0) -- (-2,0,-2) -- (1,0,-1) -- (0,2,0) -- cycle  ;
			\draw[fill=blue!30] (0,2,0) -- (2,0,2) -- (-1,0,1) -- (0,2,0) -- cycle  ;

			%---NODES---
			\node (1) at (0,2.3,0)  {$1$};
			\node (2) at (-2.3,0,-2) {$6$};
			\node (3) at (1.3,0,-1)  {$3$};
			\node (4) at (2.3,0,2) {$4$};
			\node (5) at (-1.3,0,1) {$5$};
			\node (6) at (0,-2.3,0) {$2$};

		\end{tikzpicture}
	\end{center}
	The spherical structure of this graph creates symmetry allowing for multiple minimum edge clique covers.
	In particular, this gives an example to where we lack identifiability and Theorem~\ref{thm: ident} does not apply.
\end{example}

Interestingly, \Cref{ex:nonident} illustrates that how identifiability breaks down in practice, in the sense that if the true minimum MCM graph is $\mathcal{G}$, but the wrong edge clique cover is chosen, we could learn $\mathcal{H}$ instead.
Future work, possibly incorporating interventional knowledge, to address such identifiability concerns would be of interest.

\section{Detailed comparison to related work}
\label[appendix]{app:detailedcomp}

Below we describe in detail recent related work, and make comparisons between the assumptions made there and in our work.
We focus on work with nonparametric identifiability guarantees, and make comparisons with our Theorem~\ref{thm: ident}.

First, we highlight an important distinction between our work and previous work related to confounding.
The problem of confounding, broadly construed, refers to the estimation of causal relationships \emph{between observables} (i.e., measurement variables) when there are unobserved latent variables (i.e., unmeasured confounders).
Crucially, identifying and estimating the latent variables themselves is not an issue.
This is orthogonal to our setting, where causal relationships between observables are either meaningless or not the object of interest (e.g. pixels).
By contrast, we are interested specifically in \emph{latent factor recovery}, i.e., the identification and estimation of latent structure in causal problems.
The latter is the typical setting in factor analysis, or more broadly, causal representation learning.
To summarize:
\begin{itemize}
	\item \emph{Confounding:} Learn edges $M_i\to M_j$, but not latents $L_i$
	\item \emph{Factor analysis / representation learning:} Learn latents $L_i$, influences $L_i\to M_j$, but not edges $M_i\to M_j$
\end{itemize}

With this in mind, all existing results require various functional restrictions or parametric assumptions, and generally require auxiliary information, additivity, and/or no interaction effects.
NCFA does not require these assumptions, and thus our framework applies more broadly, although it does not consider the most general forms of confounding, as in \citet{maeda2021causal} and described above.
The tradeoff is that NCFA guarantees model-free latent factor recovery, whereas papers focused on confounding impose model-specific functional assumptions to handle more general forms of confounding (i.e., to solve a different problem).
We emphasize once again that NCFA does not even need assumptions on the type of data, as it can be continuous, discrete, or mixed.

\begin{itemize}
	\item \citet{maeda2021causal} (CAM-UV) study causal additive models with unobserved variables, with a focus on causal discovery under unmeasured confounding.
	      In order to identify confounders, this paper assumes an additive noise model that is further additive over the parents, similar to CAM \citep{peters2014}, along with faithfulness-type assumptions.
	      As dscribed above, their focus on confounding is orthogonal to the present work: \citet{maeda2021causal} study causal discovery \emph{between} measurement variables where latent variables appear as nuisance confounders, whereas in latent factor models (i.e., here) the latent variables are the target of interest.
	\item \citet{yang2021causalvae} (CausalVAE) introduce CausalVAE for structured causal disentanglement.
	      Built upon the framework of auxiliary variables introduced by \citet{khemakhem2020variational}, this paper makes several assumptions that we do not require: 1) The latent SEM is linear, and 2) They observe additional signal in the form of auxiliary variables $u$ associated with the true causal concepts.
	      The latter assumption is crucial for identification.
	\item \citet{shen2022weakly} (DEAR) again use auxiliary variables (i.e., iVAE) as weak supervision for disentangled generative causal representation learning.
	      In addition to extra auxiliary information in the form of annotated labels of the ground-truth factors, a super-structure of the latent DAG is required, as the authors ``leave incorporating causal discovery from scratch to future work''.
	      Moreover, they make parametric assumptions including a linear latent SEM and a post-nonlinear model inspired by DAG-GNN \citep{yu2019dag}.
	\item \citet{ashman2022causal} (Neural-ADMG / DDECI) learn latent causal models in the form of an acyclic directed mixed graph (ADMG).
	      As with \citet{yang2021causalvae}, they make several strong identifying assumptions that we do not require: 1) Additive noise models, 2) No interaction between parents and confounders, and 3) At least two pure children for each latent.
	      In our case, only one pure child \emph{suffices}, but is not necessary, for identification (see Theorem~\ref{thm: ident} and its special case Corollary~\ref{cor:ident}).
	\item \citet{pmlr-v202-kaltenpoth23a} (No-cadillac) perform nonlinear causal discovery with latent confounders, similar to \citet{maeda2021causal}.
	      As in \citet{maeda2021causal}, the focus is on confounding (i.e., relationships between measurement variables), which are restricted by a post-nonlinear model similar to DAG-GNN \citep{yu2019dag}.
	      The latent variables are assumed to be i.i.d.
	      Gaussian, and their results impose four additional nontrivial graphical assumptions.
	      By contrast, NCFA requires two simple graphical constraints (expressed in Theorem~\ref{thm: ident}), and imposes no parametric assumptions.
	\item \citet{li2023nonlinear} (DeFuSE) also consider nonlinear causal discovery with  confounders by assuming the latent structure is completely known (and in particular, $\dim(Z)=\dim(X)$) along with Gaussian noise and Gaussian confounders.
	      They further assume an additive noise model and that each nonlinearity is of sublinear growth.
	      None of these assumptions are required by our approach, and in particular we allow for any number of latents, independent of $\dim(X)$.
\end{itemize}

\section{Evaluation Metrics}
\label[appendix]{sec:evaluation-metrics}

\paragraph{Distance measures}
Unlike classical VAE methods, NCFA has a causal discovery step in which it infers the UDG upon which the VAE in the model is based.
The structural Hamming distance (SHD) is commonly used to quantify the difference between graph structures and is simply defined to be the number of edges that appear in one graph but not the other.
SHD, however, is not applicable in our case as (i) it is defined for graphs having the same number of vertices, whereas for a fixed finite sample from a set of measurement variables, differently estimated minimum MCMs may have differing numbers of latent variables, and (ii) even if we compute, e.g., the SHD between the undirected graphs \(\mathcal{U}, \mathcal{U}'\) (which \emph{do} have the same number of vertices), this aligns poorly with intuitions about distance between their respective generating MCM graphs \(\mathcal{G}, \mathcal{G}'\).
To remedy this, we introduce the following distance:
\begin{definition}
	The \emph{shared factor distance} between biadjacency matrices \(B_1 \in \{0, 1\}^{K_1, n}\) and \(B_2 \in \{0, 1\}^{K_2, n}\), corresponding to minimum MCM graphs \(\mathcal{G}_1, \mathcal{G}_2\) each with \(n = |M|\) measurement variables and $K_i$ latents, respectively, for $i = 1,2$, is defined as \[ \textup{SFD}(B_1,B_2) \coloneqq \sum_{i\leq j} \left\lvert (B_1^\top B_1)_{ij} - (B_2^\top B_2)_{ij} \right\rvert.
	\]
\end{definition}

The SFD is a proper distance metric \citep{deza2016encyclopedia}, satisfying nonnegativity, symmetry, the triangle inequality, and (as detailed below) identity of indiscernibles.
The intuition behind this definition is that we transform a given minimum MCM structure into a weighted undirected graph, allowing it to be easily compared to other minimum MCM structures over the same set of measurement variables.
Via this transformation, the weights keep track of how many latent parents each pair of measurement variables have in common as the $(i,j)$-th entry in the matrix $B_1^\top B_1$ records the number of colliderless paths between measurement variables $i$ and $j$ in the graph.
For our graphs, two nodes admit such a path between them if and only if they are connected by a latent, and this connection adds exactly 1 to the $(i,j)$-th matrix entry.
Hence, these matrices will be equal if and only if the MeDIL models are identical (for a consistent ordering of measurement variables) \citep[Section~2.2]{deligeorgaki2022combinatorial}.
Figure~\ref{fig:sfd} provides an example comparing the SFD between minimum MCMs and the SHD between their corresponding (unweighted) undirected graphs.

\tikzstyle{VertexStyle} = []
\SetGraphUnit{2.2}
\begin{figure}[h]
	\centering
	\tikzstyle{EdgeStyle} = [->,>=stealth']
	\subfigure[\(\mathcal{G}_1\)][.33\textwidth]{
		\begin{tikzpicture}[scale=.4]
			\begin{scope}[execute at begin node=$, execute at end node=$]
				\Vertex{m_1} \EA(m_1){m_2} \EA(m_2){m_3} \EA(m_3){m_4}
				\NO(m_2){l_1} \EA(l_1){l_2}
				\Edges(l_1, m_1) \Edges(l_1, m_2) \Edges(l_1, m_3)
				\Edges(l_2, m_4) \Edges(l_2, m_2) \Edges(l_2, m_3)
			\end{scope}
		\end{tikzpicture}
	}\quad
	\subfigure[\(\mathcal{G}_2\)][.32\textwidth]{
		\begin{tikzpicture}[scale=.4]
			\begin{scope}[execute at begin node=$, execute at end node=$]
				\Vertex{m_1} \EA(m_1){m_2} \EA(m_2){m_3} \EA(m_3){m_4}
				\NO(m_1){l_1} \EA(l_1){l_2} \EA(l_2){l_3} \EA(l_3){l_4}
				\Edges(l_1, m_1) \Edges(l_1, m_2)
				\Edges(l_2, m_1) \Edges(l_2, m_3)
				\Edges(l_3, m_2) \Edges(l_3, m_4)
				\Edges(l_4, m_3) \Edges(l_4, m_4)
			\end{scope}
		\end{tikzpicture}
	}\quad
	\subfigure[\(\mathcal{G}_3\)][.33\textwidth]{
		\begin{tikzpicture}[scale=.4]
			\begin{scope}[execute at begin node=$, execute at end node=$]
				\Vertex{m_1} \EA(m_1){m_2} \EA(m_2){m_3} \EA(m_3){m_4}
				\NO(m_2){l_1} \EA(l_1){l_2}
				\Edges(l_1, m_1) \Edges(l_1, m_2) \Edges(l_1, m_3)
				\Edges(l_2, m_4) \Edges(l_2, m_3)
			\end{scope}
		\end{tikzpicture}
	}\vspace{1em}

	\tikzstyle{EdgeStyle} = []
	\subfigure[][.33\textwidth]{
		\begin{tikzpicture}[scale=.4]
			\begin{scope}[execute at begin node=$, execute at end node=$]
				\Vertex[empty]{a} \EA(a){m_4} \NO(a){m_2} \SO(a){m_3} \WE(a){m_1} \NO(m_1){U_1}
				\Edges(m_1, m_2, m_4, m_3, m_1) \Edges(m_2, m_3)
			\end{scope}
		\end{tikzpicture}
	}\quad
	\subfigure[][.32\textwidth]{
		\begin{tikzpicture}[scale=.4]
			\begin{scope}[execute at begin node=$, execute at end node=$]
				\Vertex[empty]{a} \EA(a){m_4} \NO(a){m_2} \SO(a){m_3} \WE(a){m_1} \NO(m_1){U_2}
				\Edges(m_1, m_2, m_4, m_3, m_1)
			\end{scope}
		\end{tikzpicture}
	}\quad
	\subfigure[][.33\textwidth]{
		\begin{tikzpicture}[scale=.4]
			\begin{scope}[execute at begin node=$, execute at end node=$]
				\Vertex[empty]{a} \EA(a){m_4} \NO(a){m_2} \SO(a){m_3} \WE(a){m_1} \NO(m_1){U_3}
				\Edges(m_1, m_2, m_3, m_1) \Edges(m_3, m_4)
			\end{scope}
		\end{tikzpicture}
	}\vspace{1em}

	\subfigure[][.33\textwidth]{
		\(B_1 =
		\begin{bmatrix}
			1 & 1 & 1 & 0 \\
			0 & 1 & 1 & 1
		\end{bmatrix}
		\)
	}\quad
	\subfigure[][.32\textwidth]{
		\(B_2 =
		\begin{bmatrix}
			1 & 1 & 0 & 0 \\
			1 & 0 & 1 & 0 \\
			0 & 1 & 0 & 1 \\
			0 & 0 & 1 & 1
		\end{bmatrix}
		\)
	}\quad
	\subfigure[][.33\textwidth]{
		\(B_3 =
		\begin{bmatrix}
			1 & 1 & 1 & 0 \\
			0 & 0 & 1 & 1
		\end{bmatrix}
		\)
	}\vspace{1em}

	\subfigure[]{
		\begin{tabular}{r | c c c}
			\toprule
			{}                          & {\(\mathcal{G}_1\) and \(\mathcal{G}_2\)} & {\(\mathcal{G}_1\) and \(\mathcal{G}_3\)} & {\(\mathcal{G}_2\) and \(\mathcal{G}_3\)} \\\hline
			structural Hamming distance & 1                                         & 1                                         & 2                                         \\
			shared factor distance      & 10                                        & 5                                         & 7                                         \\
			\bottomrule
		\end{tabular}
	}
	\caption{Pairwise comparison of structural Hamming distance and shared factor distance on three causal factor structures, \(\mathcal{G}_1, \mathcal{G}_2,\) and \(\mathcal{G}_3\).
		For each structure \(\mathcal{G}_i\), we also show its associated undirected graph \(U_i\) and its biadjacency matrix \(B_i\).
	}
	\label{fig:sfd}
\end{figure}

%%% Local Variables:
%%% mode: latex
%%% TeX-master: "main"
%%% End:

\end{document}